\documentclass[10pt,twocolumn,letterpaper]{article}

\usepackage{cvpr}
\usepackage{times}
\usepackage{epsfig}
\usepackage{graphicx}
\usepackage{amsmath}
\usepackage{amssymb}

\usepackage{graphicx}
\usepackage{subfigure}
\usepackage{paralist}
\usepackage{booktabs}

\usepackage[pagebackref=true,breaklinks=true,letterpaper=true,colorlinks,citecolor=blue,linkcolor=blue,bookmarks=false]{hyperref}

\cvprfinalcopy 

\def\cvprPaperID{1225} 

 \ifcvprfinal\fi
\begin{document}

\title{Target-Aware Deep Tracking}

\author{
    Xin Li$^{1}$
    \hspace{10pt}
    Chao Ma$^{2}$
    \hspace{10pt}
    Baoyuan Wu$^{3}$
    \hspace{10pt}
    Zhenyu He$^{1}$\thanks{Corresponding author.}
    \hspace{10pt}
    Ming-Hsuan Yang$^{4,5}$\\
    $^1$Harbin Institute of Technology, Shenzhen\hspace{15pt}\\
    $^2$MoE Key Lab of Artificial Intelligence, AI Institute, Shanghai Jiao Tong University\\
    $^3$Tencent AI Lab\hspace{15pt}
    $^4$University of California, Merced\hspace{15pt}
    $^5$Google Cloud AI\\
    {\tt\small\{xinlihitsz, wubaoyuan1987\}@gmail.com, chaoma@sjtu.edu.cn,}\\
    {\tt\small zhenyuhe@hit.edu.cn, mhyang@ucmerced.edu}
}

\maketitle

\begin{abstract}
Existing deep trackers mainly use convolutional neural networks pre-trained for generic object recognition task for representations.
Despite demonstrated successes for numerous vision tasks,
the contributions of using pre-trained deep features for visual
tracking are not as significant as that for object recognition.
The key issue is that in visual tracking the targets of interest can be arbitrary object class with arbitrary forms.
As such, pre-trained deep features are less effective in modeling these targets of arbitrary forms
for distinguishing them from the background.
In this paper, we propose a novel scheme to learn target-aware features, which can better recognize the targets undergoing significant appearance variations than pre-trained deep features.
To this end, we develop a regression loss and a ranking loss to guide the generation of
target-active and scale-sensitive features.
We identify the importance of each convolutional filter
according to the back-propagated gradients and
select the target-aware features based on activations for representing the targets.
The target-aware features are integrated with a Siamese matching network for visual tracking.
Extensive experimental results show that the proposed algorithm performs favorably against the state-of-the-art methods in terms of accuracy and speed.
\end{abstract}

\section{Introduction}
Visual tracking is one of the fundamental computer vision problems with a wide range of applications.
%
Given a target object specified by a bounding box in the first frame, visual tracking aims to locate the target object in the subsequent frames.
This is challenging as target objects often undergo significant appearance changes over time and may temporally leave the field of the view.
Conventional trackers prior to the advances of deep learning mainly consist of a feature extraction module and a decision-making mechanism.
The recent state-of-the-art deep trackers often use deep models pre-trained for the object recognition task to extract features, while putting more emphasis on designing effective decision-making modules.
While various decision models, such as correlation filters~\cite{KCF}, regressors~\cite{GOTURN,CREST,DRT,LSART}, and classifiers~\cite{CNN-SVM,MDNET,HEDGE}, are extensively explored, considerably less attention is paid to learning more discriminative deep features.
\begin{figure}[t]
	\includegraphics[width=0.48\textwidth]{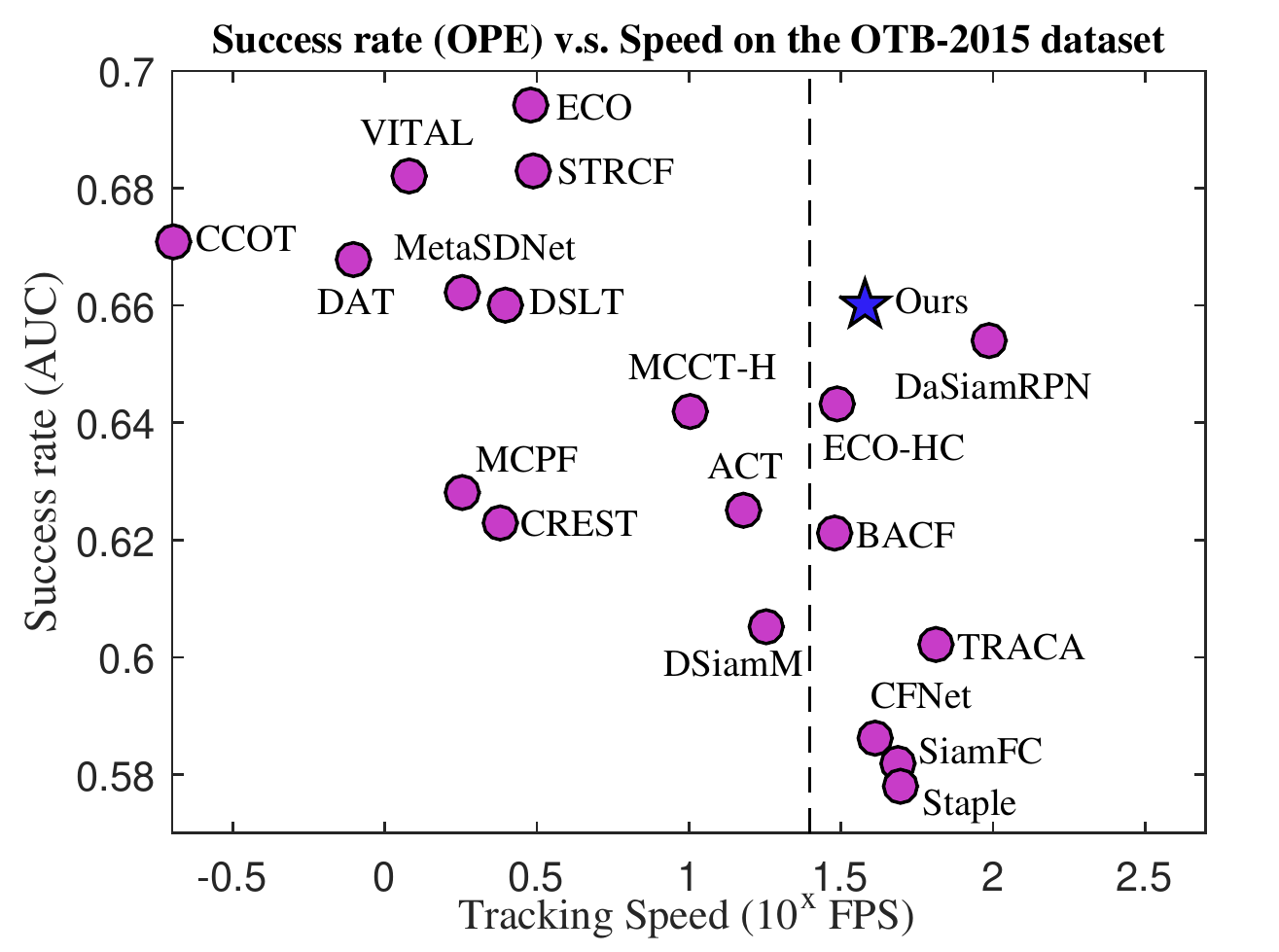}
	\caption{\textbf{Tracking accuracy vs. speed on the OTB-2015 dataset.} The horizontal and vertical coordinates correspond to tracking speed and AUC overlap ratio score, respectively. The proposed algorithm achieves a favorable performance against the state-of-the-art trackers.}
	\label{aucspeed}
\end{figure}

Despite the state-of-the-art performance of existing deep trackers, we note that the contributions of pre-trained deep features for visual tracking are not as significant as that for object recognition.
Numerous issues may arise when using pre-trained deep features as target representation.
First, a target in visual tracking can be of arbitrary forms, \eg, an object unseen in the training sets for the pre-trained models or one specific part, which does not contain the objectness information exploited for the object recognition task.
That is, pre-trained CNN models from generic images are agnostic of a target object of interest
and less effective in separating them from the background.
Second, even if target objects appear in the training set for pre-trained models, deep features taken from the last convolutional layers often retain only high-level visual information that is less effective for precise localization or scale estimation.
Third, state-of-the-art deep trackers~\cite{MDNET,CREST,VITAL} require high computational loads as deep features from pre-trained models are high-dimensional (see Figure~\ref{aucspeed}).
%
%
To narrow this gap, it is of great importance to exploit deep features pertaining specifically
to target objects for visual tracking.

\begin{figure}[t]
    \centering
    \subfigure[Distributions of intra-class targets (pedestrian).]{
        \begin{minipage}[b]{1\linewidth}
             \includegraphics[width=1\linewidth]{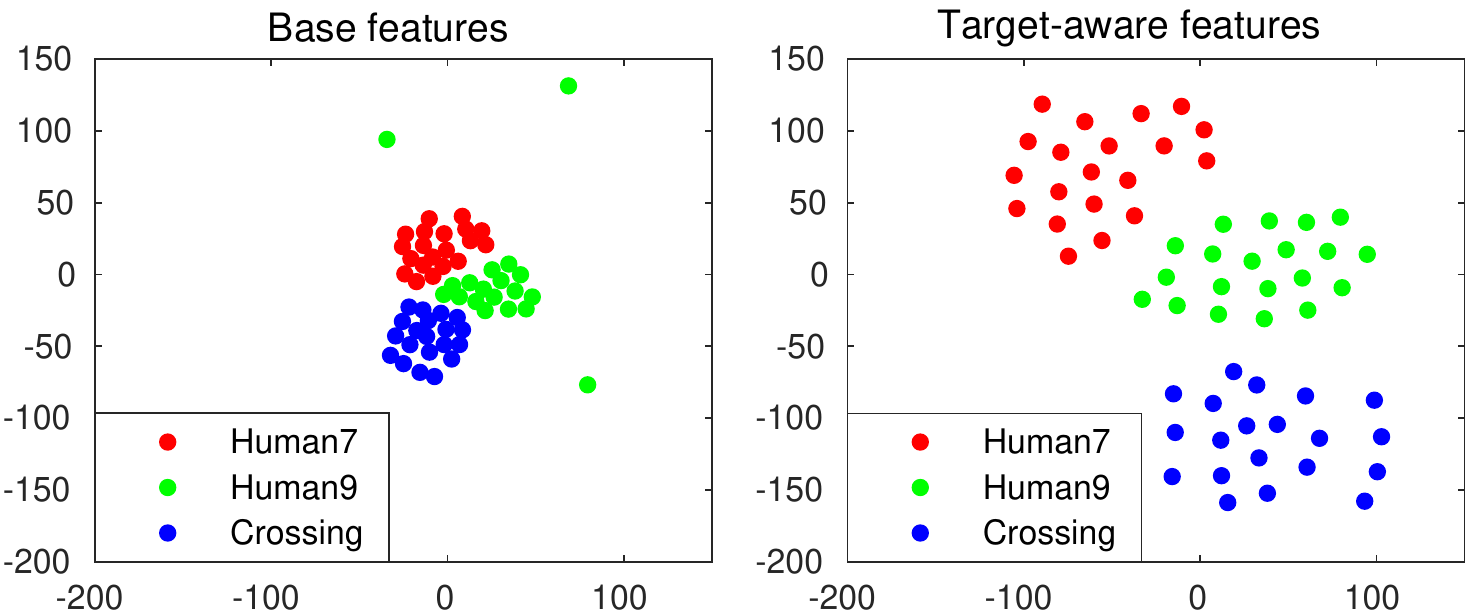}
    \end{minipage}
    }
    \subfigure[Distributions of inter-class targets.]{
        \begin{minipage}[b]{1\linewidth}
             \includegraphics[width=1\linewidth]{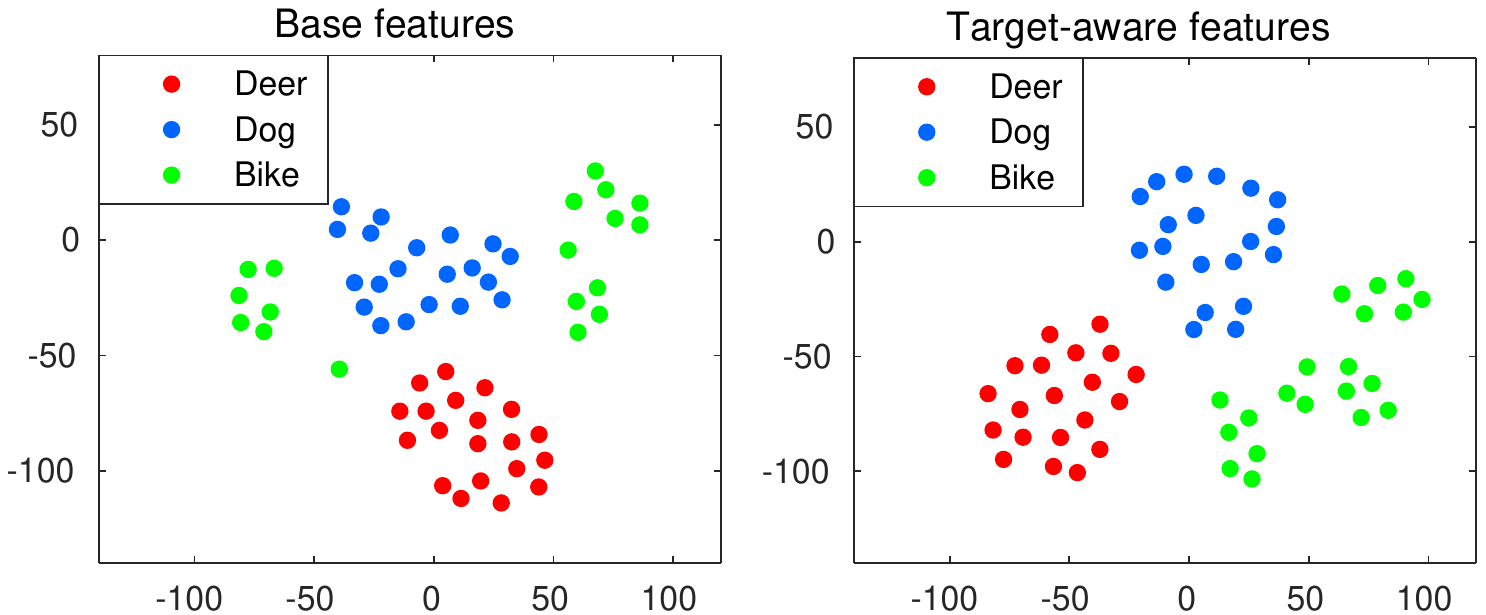}
        \end{minipage}
    }
    \caption{\textbf{Pre-trained classification CNNs features and target-aware features using the t-SNE method.}
 In this example, we randomly select 20 frames from each video.
 Each point in the figure denotes a target in one frame.
(a) All points belong to the pedestrian class but in different videos.
The target-aware features are more sensitive to intra-class differences for each video,
which are crucial for distinguishing the target from distractors.
(b) Points of different colors belong to different object classes.
The target-aware features separate objects of different categories more effectively, which can be used to remove unrelated filters and retaining target-active filters.
}
\label{fig:gap}
\end{figure}

To address the above-mentioned issues, we propose to learn target-aware deep features.
Our work is motivated based on the following observations.
The gradients obtained through back-propagating a classification neural network indicate
class-specific saliency well~\cite{GCAM}.
With the use of global average pooling, the gradients generated by a convolutional filter can determine the importance of a filter for representing target objects.
To select the most effective convolutional filters, we design two types of objective losses to perform back-propagation on top of a pre-trained deep model in the first frame.
We use a hinge loss to regress pre-trained deep features to soft labels generated by a Gaussian function and use the gradients to select the target-active convolutional filters.
We use a ranking loss with pair-wise distance to search for the scale-aware convolutional filters.
The activations of the selected most important filters are the target-aware features in this work.
Figure~\ref{fig:gap} shows the target-aware features and original deep features using the t-SNE method~\cite{t-SNE}.
Note that the target-aware deep features are more effective in separating different target objects with a same semantic label than the pre-trained deep features, which are agnostic of the objectness of the targets.
As we exploit a small set of convolutional filters to generate target-aware features, the feature number is significantly reduced, which can reduce computational loads.

We integrate the proposed target-aware features with a Siamese matching network~\cite{SIAMESEFC} for visual tracking.
%
%
We evaluate the proposed tracker on five benchmark datasets including OTB-2013~\cite{OTB2013}, OTB-2015~\cite{OTB2015}, VOT-2015~\cite{VOT2015,VOTTPAMI}, VOT-2016~\cite{VOT2016}, and Temple Color-128~\cite{TC128}.
Extensive experiments with ablation studies demonstrate that
the proposed target-aware features are more effective than those from pre-trained models for the Siamese trackers in terms of accuracy and tracking speed.

The main contributions of this work are summarized as follows:
\begin{compactitem}
\item We propose to learn target-aware deep features for visual tracking.
We develop a regression loss and a ranking loss for selecting the most effective convolutional filters to generate target-aware features.
We narrow the gap between the pre-trained deep models and target objects of arbitrary forms for  visual tracking.
\item We integrate the target-aware features with a Siamese matching network for visual tracking.
The target-aware features with reduced number of features can accelerate Siamese trackers as well.
\item We evaluate the proposed method extensively on five benchmark datasets.
 We show that the Siamese tracker with the proposed target-aware features performs well against the state-of-the-art methods in terms of effectiveness and efficiency.
\end{compactitem}

\section{Related Work}
\label{related_work}
Visual tracking has been an active research topic in the literature.
In the following, we mainly discuss the representative deep trackers and
related issues on the gradient-based deep models.
%

\vspace{2mm}
\noindent\textbf{Deep trackers.}
One notable issue of applying deep learning models to visual tracking is that there are limited training samples and only the ground truth visual appearance
of the target object in the first frame is available.
On one hand, most existing deep trackers
use deep models pre-trained for the object classification task for feature representations.
Several trackers~\cite{HCF,VTWFCN} exploit the complementary characteristics of shallow and deep layer features to enable the abilities of robustness and accuracy.
Deep features from multiple layers have also integrated for visual tracking~\cite{CCOT,HEDGE,ECO,UPDT}.
However, the combination of pre-trained deep features may not always bring performance gains, due to issues of unseen targets, incompatible resolutions, and increasing dimensions, as demonstrated by Bhat \etal~\cite{UPDT}.
On the other hand, numerous trackers~\cite{CNN-SVM,ACFT,CACFT,BACF,CREST,ADNET,BRANCHOUT} are developed by improving the decision models
including support vector machines, correlation filters, deep classifiers, and deep regressors.
Nam and Han~\cite{MDNET} propose a multi-domain deep classifier combined with the hard negative mining, bounding box regression, and online sample collection modules for visual tracking.
The VITAL tracker~\cite{VITAL} exploits adversarial learning to generate effective samples and leverages the class imbalance with a cost-sensitive loss.
However, these models may drift from target object in the presence of noisy updates and require high computational loads, which is caused by the limited online training samples to a large extent.

To exploit datasets with general objects for tracking, numerous Siamese based trackers~\cite{SIAMESEFC,SINT,DSIAM, SIAMRPN,GOTURN} cast tracking as a matching problem and learn a similarity measurement network.
Tracking is carried out by comparing the features of the initial target template and search regions in the current frame.
A number of trackers~\cite{RASNET,FLOWT,TWOFOLD} have since been developed by introducing attention mechanisms for better matching between templates and search regions.
Although these Siamese frameworks are pre-trained on large video datasets,
the pair-wise training sample only tells whether the two samples belong to the same target or not without category information.
That is, the Siamese trackers do not fully exploit semantic and objectness information pertaining to specific target objects.
In this work, we select the most discriminative and scale-sensitive convolutional filters from a pre-trained CNN to generate target-aware deep features.
The proposed features enhance the discriminative representation strength of the targets regarding semantics and objectness, which
facilitate the Siamese tracking framework to perform well against the state-of-the-art methods
in terms of robustness and accuracy.

\vspace{2mm}
\noindent\textbf{Gradient-based deep models.}
Several gradient-based models~\cite{CAM,GCAM} are developed to determine the importance of each channel of CNN features in describing a specific object class.
The GCAM model~\cite{CAM} generates a class-active map by computing a weighted sum along the feature channels based on the observation that the gradient at each input pixel indicates the corresponding importance belonging to given class labeling.
The weight of a feature channel is computed by globally average pooling of all the gradients in this channel.
Unlike these gradient-based models using classification losses, we specifically design a regression loss and a ranking loss for the tracking task to identify
which convolutional filters are active to describe targets and sensitive to scale changes.

\section{Target-Aware Features}
In this section, we present how to learn target-aware features for visual tracking.
We first analyze the gap between the features from pre-trained classification deep models and effective representations for visual tracking.
Then, we present the target-aware feature model including a discriminative feature generation model and a scale-sensitive feature generation component based on the gradients of regression and ranking losses.
%
\begin{figure*}
    \centering
    \includegraphics[width=0.9\textwidth]{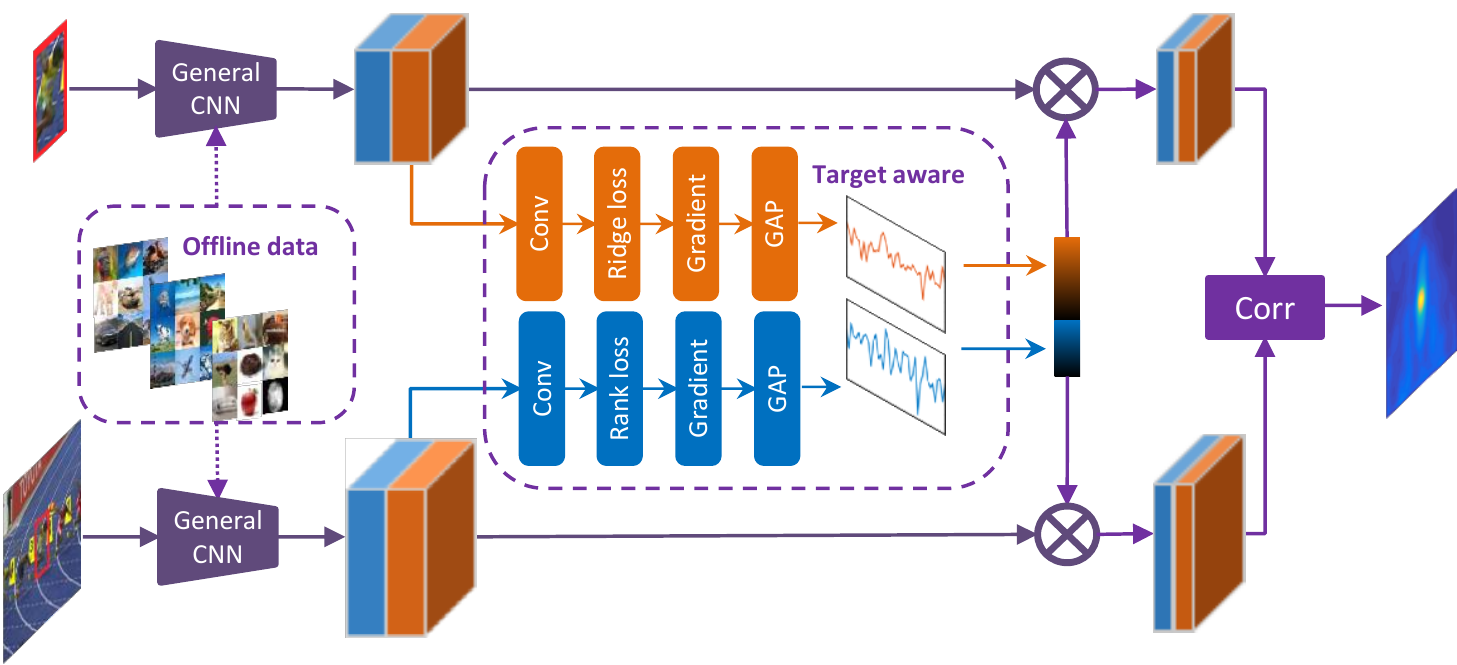}
    \caption{\textbf{Framework of the proposed algorithm.} This framework consists of a general CNN feature backbone network, a target-aware model, and a correlation matching module.
    The target-aware model, constructed with a regression loss part (\ie, Ridge loss) and a ranking loss part, selects the target-aware filters with target-active and scale-sensitive information from the pre-trained CNNs for object recognition.
  The correlation matching module computes the similarity score between the template and the search region.
The maximum of the score map indicates the target position.}
    \label{fig:framework}
\end{figure*}

\subsection{Features of pre-trained CNNs}
The gap between the features effective for generic visual recognition and
object-specific tracking is caused by the following issues.
First, the pre-trained CNN features are agnostic of the semantic and objectness information of the target, which most likely does not appear in the offline training data.
Different from other vision tasks (\eg, classification, detection, and segmentation), where the class categories for training and testing are pre-defined and consistent, online visual tracking needs to deal targets of any object labels.
Second, the pre-trained CNNs focus on increasing inter-class differences and the extracted deep features are insensitive to intra-class variations.
As such, these features are less effective for trackers to accurately estimate scale changes and distinguish the targets from distractors with the same class label.
Third, the pre-trained deep features are sparsely activated by each category label (\ie, inter-class difference are mainly related to a few feature channels)
especially in a deeper convolutional network.
When applied to the tracking task, only a few convolutional filters are active in describing the target.
 A large portion of the convolutional filters contain redundancy and irrelevant information, which leads to high computational loads and over-fitting.
Figure~\ref{fig:gap} shows the distributions of the pre-trained deep features and the proposed target-aware features using the t-SNE method~\cite{t-SNE}.

Several methods on interpretation of neural networks demonstrate that the importance of convolutional filters on capturing the category-level object information
can be computed through the corresponding gradients~\cite{CAM,GCAM}.
Based on the gradient-based guidance,
we construct a target-aware feature model with losses designed specifically for visual tracking.
Given a pre-trained CNN feature extractor with the output feature space $\chi$,
a subspace $\chi '$ can be generated based on the channel importance $\Delta$ as
\begin{equation}
    \chi ' = \varphi(\chi ; \Delta),
    \label{eq:mapping}
\end{equation}
where $\varphi$ is a mapping function selecting the most important channels.
The importance of the $i$-th channel $\Delta_i$ is computed by
\begin{equation}
    \Delta_i =G_{AP}(\frac{\partial L}{\partial z_i}),
    \label{eq:weights}
\end{equation}
where $G_{AP}(\cdot)$ denotes the global average pooling function, $L$ is the designed loss,
and $z_i$ indicates the output feature of the $i$-th filter.
For visual tracking, we exploit the gradients of a regression loss (Section~\ref{sec:RegressionLoss}) and a ranking loss (Section~\ref{sec:RankLoss})
to extract target-aware features.

\begin{figure*}[!htb]
	\centering
    \includegraphics[width=0.98\linewidth]{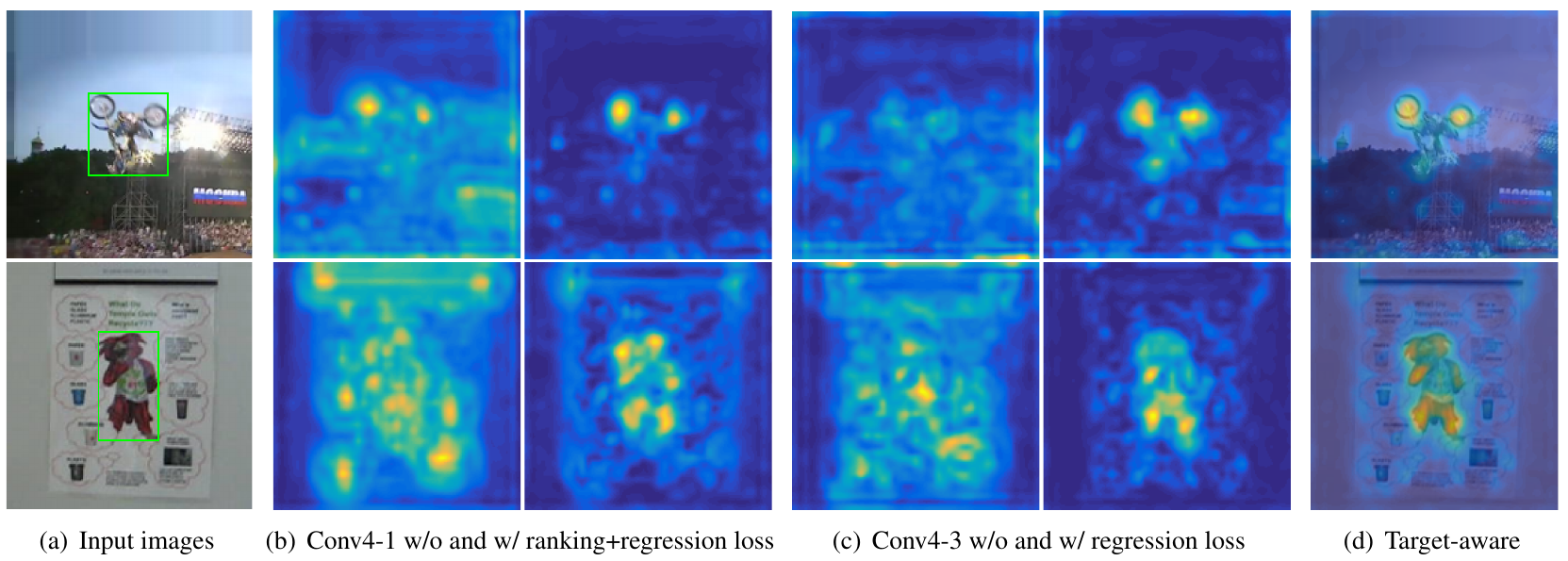}
	\caption {\textbf{Visualization of the original and the learned target-aware features.} The visualized images are generated by averaging all channels.
From left to right on each row are the input images, pre-trained deep features (\textit{Conv4-1}) without and with ranking and regression losses for learning scale-sensitive features, pre-trained deep features (\textit{Conv4-3}) without and with a regression loss for learning objectness-sensitive features, and the overall target-aware deep features.
Notice that the original pre-trained features are not effective in describing the targets, while the target-aware features can readily separate the targets from the background.}
	\label{fig:feature-maps}
\end{figure*}

\subsection{Target-Active Features via Regression}
\label{sec:RegressionLoss}
In a pre-trained classification network, each convolutional filter captures a specific feature pattern and all the filters construct a feature space containing different objectness priors.
A trained network recognizes a specific object category mainly based on a subset of these filters.
For the visual tracking task, we can obtain the filters with objectness information pertaining to the target by identifying those active to the target area while inactive to the backgrounds.
To this end, we regress all the samples $X_{i,j}$ in an image patch aligned with the target center to a Gaussian label map $Y(i,j)=e^{-\frac{i^2+j^2}{2\sigma ^2}}$, where $(i,j)$ is the offset against the target and $\sigma$ is the kernel width.
%
For computational efficiency, we formulate the problem as the ridge regression loss,
\begin{equation}
    L_{reg} = \|Y(i,j) - W*X_{i,j} \|^2 + \lambda\|W\|^2 ,
    \label{eq:regressionloss}
\end{equation}
where $*$ denotes the convolution operation and $W$ is the regressor weight.
The importance of each filter can be computed based on its contribution to fitting the label map, \ie, the derivation of $L_{reg}$ with respect to the input feature $X_{in}$.
With the chain rule and Eq.~\ref{eq:regressionloss}, the gradient of the regression loss is computed by
\begin{equation}
\begin{split}
    \frac{\partial L_{reg}}{\partial X_{in}}& =\sum_{i,j} \frac{\partial L_{reg}}{\partial X_o(i,j)} \times \frac{\partial X_o(i,j)}{\partial X_{in}(i,j)} \\
&    = \sum_{i,j}2(Y(i,j)-X_o(i,j))\times W,
\end{split}
\end{equation}
where $X_o$ is the output prediction.
With the gradient of the regression loss and Eq.~\ref{eq:weights}, we find the target-active filters that are able to discriminate the target from the background.
The generated features have the following merits compared to the pre-trained deep features.
We select a portion of target-specific filters to generate discriminative deep features.
This not only alleviates the model over-fitting issue but also reduces the number of features.
The target-aware features are effective for representing an arbitrary target or an unsee object in the training set.
Figure~\ref{fig:feature-maps}(c) visually compares the deep features learned with and without regression-loss by averaging all channels.

\subsection{Scale-Sensitive Features via Ranking}
\label{sec:RankLoss}
To generate scale-sensitive features, we need to find the filters that are most active to the target scale changes.
The exact scale of the target is hard to compute as target presentation is not continuous, but we can get the closest scale with a model that tells which one has a closer size of a paired sample.
As such, we formulate the problem as a ranking model and rank the training sample whose size is closer to the target size higher.
The gradients of the ranking loss indicate the importance of the filters to be sensitive to scale changes.
%
For ease of implementation, we exploit a smooth approximated ranking loss~\cite{LSEP} defined by
\begin{align}
    L_{rank} = \log\big(1+\sum_{(x_i,x_j)\in\Omega}\exp{(f(x_i)-f(x_j))}\big),
    \label{eq:rankingloss}
\end{align}
where $(x_i$, $x_j)$ is a pair-wise training sample and the size of $x_j$ is closer to the target size comparing with $x_i$, and $f(x;w)$ is the prediction model.
In addition, $\Omega$ is the set of training pairs.
The derivation of $L_{rank}$ with respect to $f(x)$ is computed as~\cite{LSEP}:
\begin{align}
    \frac{\partial L_{rank}}{\partial{f(x)}} = -\frac{1}{L_{rank}}\sum_\Omega \Delta \mathbf{z}_{i,j} \exp({-f(x)\Delta \mathbf{z}_{i,j}}),
    \label{eq:grad_ranking}
\end{align}
where $\Delta \mathbf{z}_{i,j}=\mathbf{z}_{i}-\mathbf{z}_j$ and $\mathbf{z}_i$ is a one-hot vector with the $i$-th element being 1 while others being 0.
By back-propagation, the gradients of ranking loss with respect to the features can be computed by
\begin{align}
     \frac{\partial L_{rank}}{\partial x_{in}}= \frac{\partial L_{rank}}{\partial x_o} \times \frac{\partial x_o}{\partial x_{in}}  = \frac{\partial L_{rank}}{\partial{f(x_{in})}}  \times W,
\end{align}
where $W$ is the filter weights of the convolutional layer.
With the above gradients of the ranking loss and Eq.~\ref{eq:weights}, we find the filters that are sensitive to scale changes.
Considering we only need the scale-sensitive features of the target object, we combine the regression and ranking losses to find the filters that are both active to the target and sensitive to scale changes.
Figure~\ref{fig:feature-maps}(b) visually compares deep features generated with and without the proposed model by averaging all channels.

\section{Tracking Process}
Figure~\ref{fig:framework} shows the overall framework of the proposed tracker.
We integrate the target-aware feature generation model with the Siamese framework due to the following two reasons.
First, the Siamese framework is concise and efficient as it performs tracking by comparing the features of the target and the search region.
Second, the Siamese framework can highlight the effectiveness of the proposed feature model, as its performance solely hinges on the effectiveness of the features.
We briefly introduce the tracking process with the following modules.

\vspace{2mm}
\noindent\textbf{Tracker initialization.}
The proposed tracking framework comprises a pre-trained feature extractor, the target-aware feature module, and a Siamese matching module.
The pre-trained feature extractor is offline trained on the classification task and the target-aware part is only trained in the first frame.
In initial training, the regression loss and the ranking loss parts are trained separately and we compute the gradients from each loss once the networks are converged.
With the gradients, the feature generation model selects a fixed number of the filters with the highest importance scores from the pre-trained CNNs.
The final target-aware features are obtained by stacking these two types of feature filters.
Considering the scalar difference, these two types of features are re-scaled by dividing their maximal channel summation (summation of all the values in one channel).

\vspace{2mm}
\noindent\textbf{Online detection.}
At the inference stage, we directly compute the similarity scores between the initial target and the search region  in the current frame using the target-aware features.
This is achieved by a convolution operation (\ie, the correlation layer in the Siamese framework) and outputs a response map.
The value in the response map indicates the confidence of its corresponding position to be the real target.
Given the initial target $x_1$, and the search region in the current frame $z_t$, the predicted target position in frame $t$ is computed as
\begin{equation}
    \hat{p} = \arg\max_p ~\chi'(x_1) * \chi'(z_t),
\end{equation}
where * denotes the convolution operation.

\vspace{2mm}
\noindent \textbf{Scale evaluation.}
To evaluate the scale change of the target, we fix the size of the template and re-scale the feature map of the search region in the current frame to smaller, larger, and fixed ones.
During tracking, all these three feature maps are compared with the target template.
The scale evaluation is performed by finding the score map containing the highest response.
%

\section{Experimental Results}
In this section, we first introduce the implementation details of the proposed tracker.
Then, we evaluate the proposed algorithm on five benchmark datasets and compare it with the state-of-the-art methods.
In addition, we conduct ablation studies to analyze the effectiveness of each module.
Source code and more results can be found at the \href{https://xinli-zn.github.io/TADT-project-page/}{project page}.
\subsection{Implementation Details}
We implement the proposed tracker in Matlab with the MatConvNet toolbox~\cite{MATCONV} on a PC with 32G memory, an i7 3.6GHz CPU, and a GTX-1080 GPU. The average tracking speed is 33.7 FPS.
We use the VGG-16 model~\cite{VGG16} as the base network.
To maintain more fine-grained spatial details, we use the activation outputs of the \textit{Conv4-3} and \textit{Conv4-1} layers as the base deep features.
In the initial training, the convergence loss threshold is set to 0.02 and the maximum iteration number is 50.
We select the top 250 important filters from the \textit{Conv4-3} layer for learning target-active features and select the top 80 important filters from the \textit{Conv4-1} layers for learning scale-sensitive features.
For the Siamese framework, we use the initial target as the template and crop the search region with 3 times of the target size from the current frame.
We resize the target template into a proper size if it is too large or small.
For the scale evaluation, we generate a proposal pyramid with three scales, \ie, 45/47, 1, and 45/43 times of the previous target size. We set the corresponding changing penalties to the pyramid to 0.990, 1, and 1.005.
%
%

\begin{table}
\small
\renewcommand\arraystretch{1.1}
\centering
\caption{\label{tab:OTBrs} \textbf{Experimental results on the OTB datasets.} The AUC scores on the OTB-2013 and OTB-2015 datasets are presented. The notation * denotes the running speed is reported by the authors as the source code is not available. From top to bottom, the trackers are broadly categorized into three classes: correlation filters based trackers, non-real-time deep trackers, and real-time deep trackers.}
\resizebox{\linewidth}{!}{
\begin{tabular}{ l c c c c c}
\toprule
Tracker             & OTB-2013 & OTB-2015   & Real-time & FPS       \\
\midrule
BACF~\cite{BACF}    & 0.657 & 0.621     &  Y    &30           \\
MCPF~\cite{MCPF}               & 0.677 & 0.628     &  N    &1.8           \\
MCCT-H~\cite{MCCT}  & 0.664 & 0.642     &  N    &10          \\
CCOT~\cite{CCOT}              & 0.672 & 0.671     &  N    &0.2               \\
STRCF~\cite{STRCF}  & 0.683 & 0.683     &  N     &3.1            \\
ECO~\cite{ECO}      & 0.702 & 0.694     &  N     &3.1              \\
DRT~\cite{DRT}      & 0.720 & 0.699     &  N     &1.0*             \\
\midrule
DSiamM~\cite{DSIAM}    & 0.656 & 0.605  &  N     &18              \\
ACT~\cite{ACT}         & 0.657 & 0.625  &  N     &15                \\
CREST~\cite{CREST}  & 0.673 & 0.623     &  N     &2.4            \\
FlowT~\cite{FLOWT}  & 0.689 & 0.655     &  N     &12*	      \\
DSLT~\cite{DSLT}    & 0.683 & 0.660     &  N     &2.5         \\
DAT~\cite{DAT}      & 0.704 & 0.668     &  N     &0.79          \\
LSART~\cite{LSART}  & 0.677 & 0.672     &  N     &1.0*           \\
MDNet~\cite{MDNET}  & 0.708 & 0.678     &  N     &1.1          \\
VITAL~\cite{VITAL}  & 0.710 & 0.682     &  N     &1.2            \\
\midrule
SiamRPN~\cite{SIAMRPN} &  0.658 & 0.637 &  Y     &71*               \\
RASNet~\cite{RASNET}   & 0.670 & 0.642  &  Y     &83*	          \\
SA-Siam~\cite{TWOFOLD} & 0.676 & 0.656  &  Y     &50* 	      \\
CFNet~\cite{CFNet}     &  0.611 &0.586  &  Y     &41               \\
SiamFC~\cite{SIAMESEFC}&  0.607 & 0.582 &  Y     &49              \\
TRACA~\cite{TRACA}     & 0.652 & 0.602  &  Y     &65            \\
DaSiamRPN~\cite{DaSiamRPN}&  0.668 & 0.654 &  Y  &97               \\
\textbf{Ours}   & \textbf{0.680} & \textbf{0.660} &  \textbf{Y} & \textbf{33.7}\\
\bottomrule
\end{tabular}}
\end{table}

\subsection{Overall Performance}
We evaluate the proposed algorithm on five benchmark datasets, including OTB-2013, OTB-2015, VOT-2015, VOT-2016, and Temple color-128.
The proposed algorithm is compared with the state-of-the-art trackers, including the correlation filters based trackers, such as SRDCF~\cite{SRDCF}, Staple~\cite{staple}, MCPF~\cite{MCPF}, CCOT~\cite{CCOT}, ECO~\cite{ECO}, BACF~\cite{BACF}, DRT~\cite{DRT}, STRCF~\cite{STRCF}, and MCCT-H~\cite{MCCT}; the non-real-time deep trackers such as MDNet~\cite{MDNET}, CREST~\cite{CREST}, LSART~\cite{LSART}, FlowT~\cite{FLOWT}, DSLT~\cite{DSLT}, MetaSDNet~\cite{Meta-tracker}, VITAL~\cite{VITAL}, and DAT~\cite{DAT}; and the real-time deep trackers such as ACT~\cite{ACT}, TRACA~\cite{TRACA}, SiamFC~\cite{SIAMESEFC}, CFNet~\cite{CFNet}, DSiamM~\cite{DSIAM}, RASNet~\cite{RASNET}, SA-Siam~\cite{TWOFOLD}, SiamRPN~\cite{SIAMRPN}, and DaSiamRPN~\cite{DaSiamRPN}.
In the following, we will present the results and analyses on each dataset.

\vspace{2mm}
\noindent\textbf{OTB dataset.}
The OTB-2013 dataset with 50 sequences and the extended OTB-2015 dataset with additional 50 sequences are two widely used tracking benchmarks.
The sequences in the OTB datasets are with a wide variety of tracking challenging, such as illumination variation, scale variation, deformation, occlusion, fast motion, rotation, and background clutters.
The OTB benchmark adopts Center Location Error (CLE) and Overlap Ratio (OR) as the base metrics~\cite{OTB2013}.
Based on CLE and OR, the precision and success plots are used to evaluate the overall tracking performance.
The precision plot measures the percentage of frames whose CLE is within a given threshold, which is usually set to 20 pixels.
The success plot computes the percentage of the successful frames whose OR is larger than a given threshold.
The area under the curve (AUC) of the success plot is mainly used to rank tracking algorithms.

\begin{figure}
    \centering
      \subfigure[Results on the OTB-2013 dataset]{
    \begin{minipage}{1\linewidth}
         \includegraphics[width=0.48\linewidth]{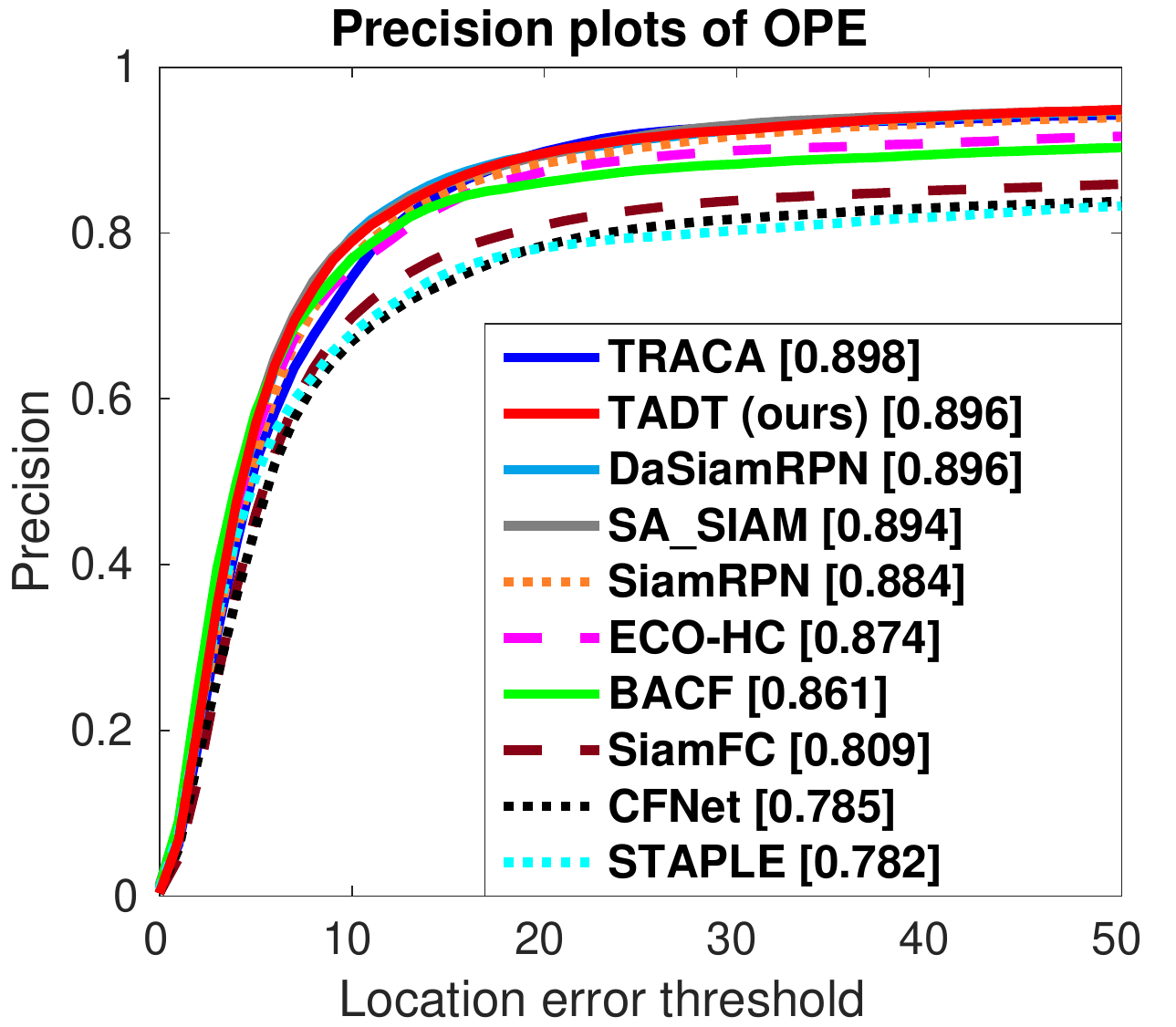}
         \includegraphics[width=0.48\linewidth]{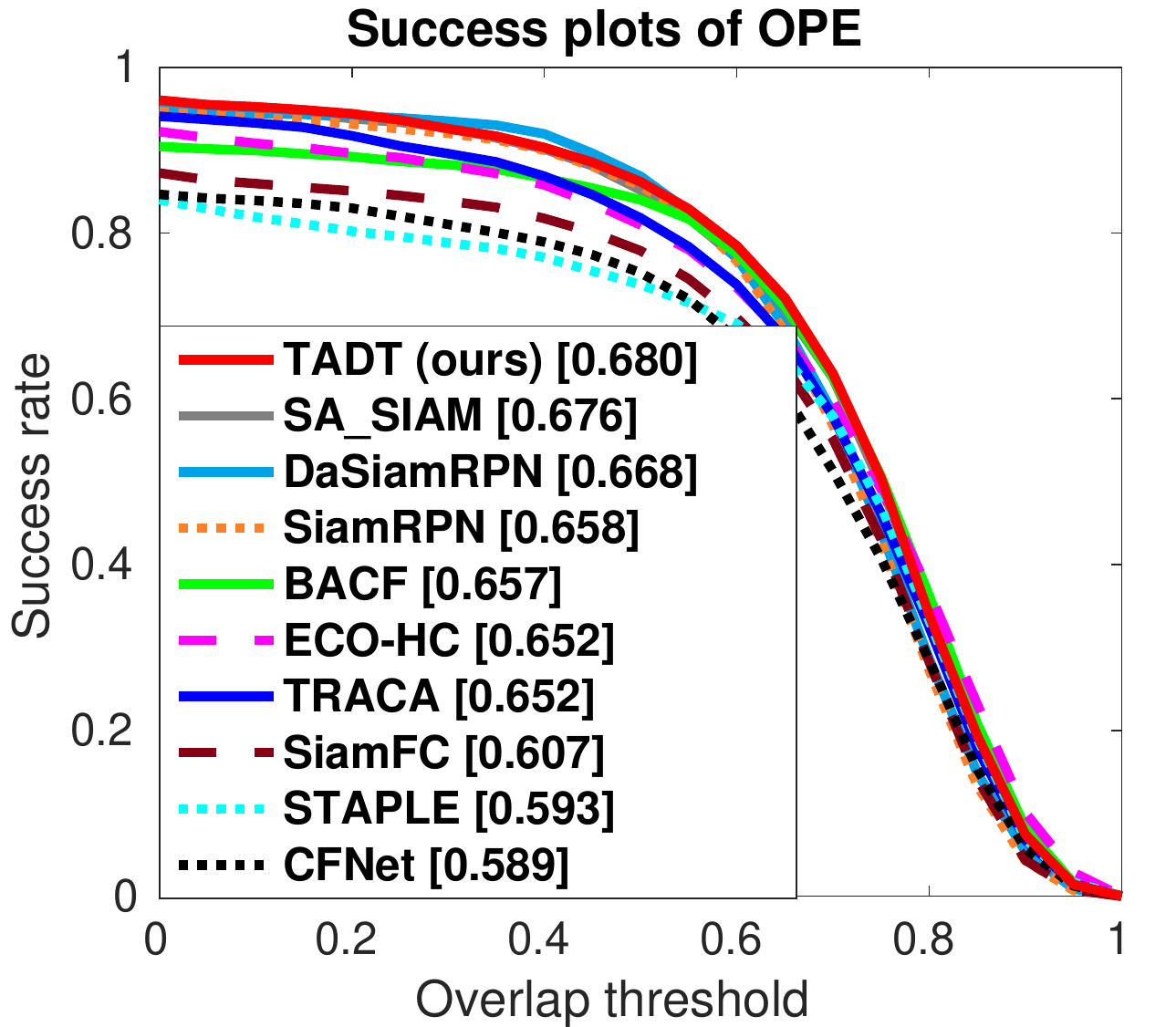}
    \end{minipage}
    }
    \subfigure[Results on the OTB-2015 dataset]{
    \begin{minipage}{1\linewidth}
         \includegraphics[width=0.48\linewidth]{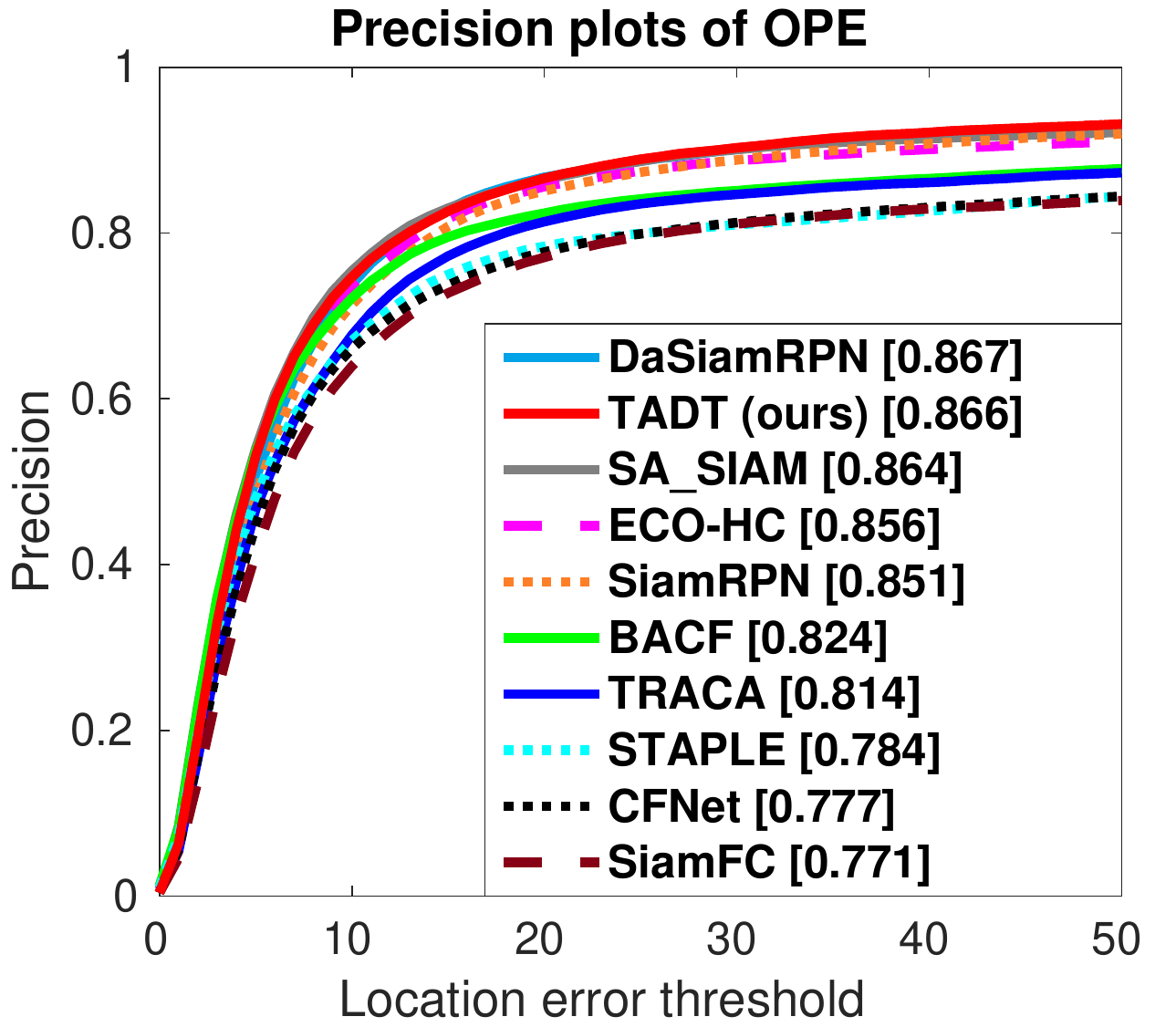}
         \includegraphics[width=0.48\linewidth]{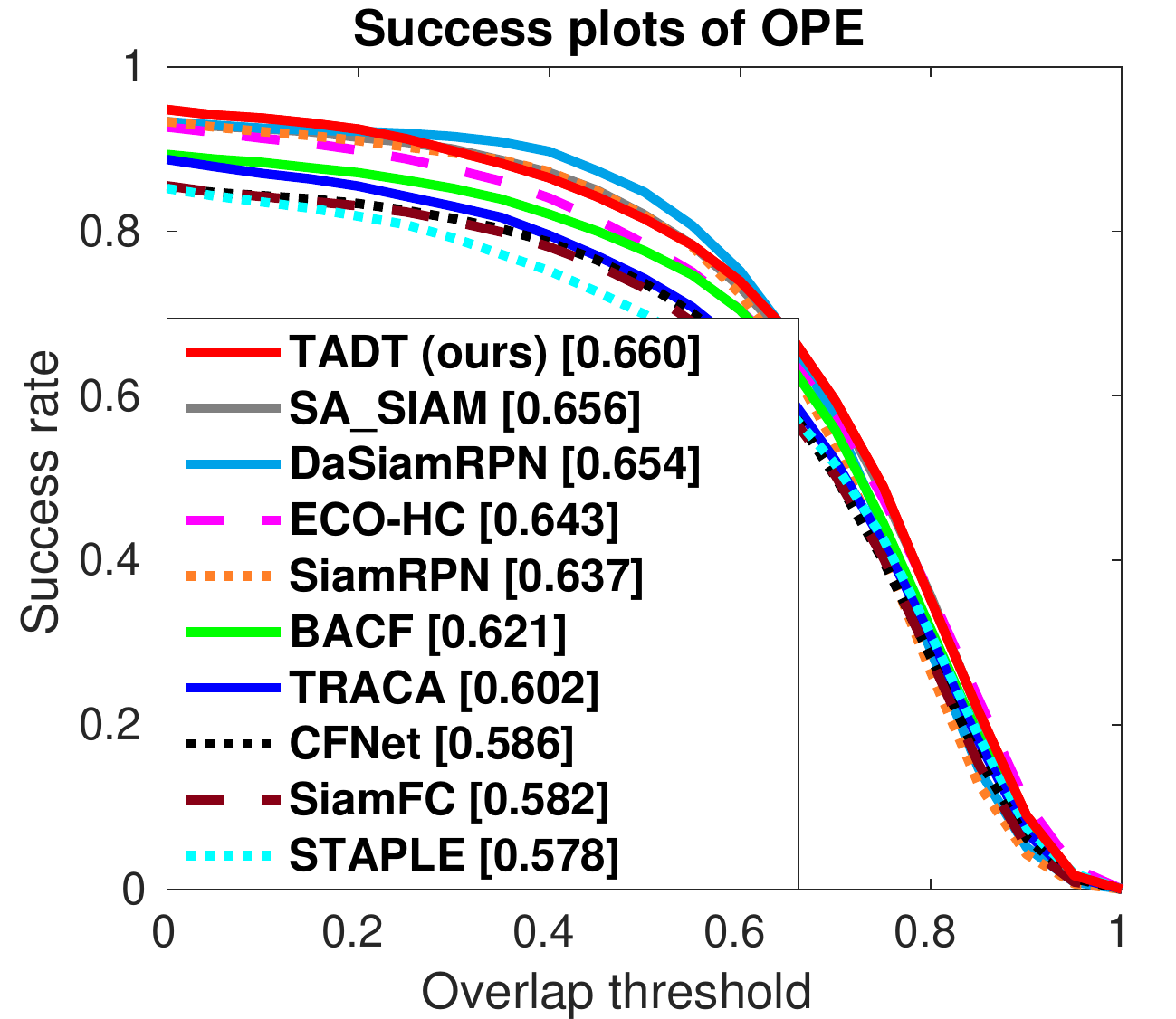}
    \end{minipage}
    }
    \caption{\textbf{Success and precision plots on the OTB-2013 and OTB-2015 datasets.}}
    \label{fig:OTBrs}
\end{figure}

Table~\ref{tab:OTBrs} shows the AUC score and the running speed of the three categories of trackers on the OTB-2013 and OTB-2015 datasets.
In the group of real-time trackers, the proposed algorithm achieves the best performance on both the OTB-2013 dataset (AUC score: 0.680) and the OTB-2015 dataset (AUC score: 0.660).
Compared with the state-of-the-art Siamese trackers with offline training, the proposed algorithm achieves the best performance on the OTB-2015 dataset.
This is because the proposed target-aware deep features best exploits the objectness and semantic information of the targets and are robust to their appearance variations as well as scale changes.
The correlation filters based trackers (DRT and ECO) achieve top performance among all the compared trackers due to the benefits from the multi-feature fusion and online updating schemes.
Non-real-time deep trackers all achieve good AUC scores. However, they suffer from time-consuming online training and model overfitting.
Equipped with the concise Siamese framework and a small set of deep features, the proposed algorithm achieves a real-time tracking speed (33.7 FPS).
This demonstrates the effectiveness of the proposed target-aware features, as the performance of the Siamese tracking framework solely hinges on the discriminative power of features.
Figure~\ref{fig:OTBrs} shows the favorable performance of the proposed tracker against the state-of-the-art real-time trackers.
For concise representation, we only show the real-time trackers ($\geq$25 FPS) in this figure, and the complete results of other trackers can be found in Table~\ref{tab:OTBrs}.
\begin{table}[!tb]
\small
\renewcommand\arraystretch{1.1}
    \caption{\textbf{Experimental results on the VOT-2015 dataset.} The notation (*) indicates the number is reported by the authors.}
    \label{tab:VOT2015}
    \centering
    \begin{tabular}{lccccc}
    \toprule
    Tracker                     & EAO $\uparrow$& Accuracy $\uparrow$& Failure$\downarrow$ & FPS \\
    \midrule
    SiamFC~\cite{SIAMESEFC}     & 0.292  & 0.54    &  1.42    & 49\\
    Staple~\cite{staple}        & 0.30    & 0.57    & 1.39     & 50\\
    SA-Siam~\cite{TWOFOLD}      & 0.31    & 0.59   & 1.26      & 50*\\
    EBT~\cite{EBT}              & 0.313  & 0.45     & 1.02    & 4.4*\\
    DeepSRDCF~\cite{DEEP-SRDCF} & 0.318 & 0.56     & 1.0     & 1*\\
    FlowT ~\cite{FLOWT}         & 0.341 & 0.57   & 0.95      &  12*\\
    \textbf{Ours}         & \textbf{0.327} & \textbf{0.59} & \textbf{1.09}   & \textbf{33.7}\\
    \bottomrule
    \end{tabular}
\end{table}

\begin{table}[!tb]
\small
\renewcommand\arraystretch{1.1} 
    \caption{\textbf{Experimental results on the VOT-2016 dataset.} The notation (*) indicates the number is reported by the authors.}
    \label{tab:VOT2016}
    \centering
    \begin{tabular}{lccccc}
    \toprule
    Tracker                 & EAO $\uparrow$& Accuracy $\uparrow$& Failure$\downarrow$ & FPS \\
    \midrule
    SA-Siam~\cite{TWOFOLD}  & 0.291    & 0.54   & 1.08      & 50*\\
    EBT~\cite{EBT}          & 0.291  & 0.47     & 0.9     & 4.4*\\
    Staple~\cite{staple}    & 0.295    & 0.54    & 1.2     & 50\\
    C-COT~\cite{CCOT}       & 0.331     & 0.53  & 0.85         & 0.3\\
    \textbf{Ours}      &\textbf{0.299} & \textbf{0.55} & \textbf{1.17}  & \textbf{33.7}\\
    \bottomrule
    \end{tabular}
\end{table}

\vspace{2mm}
\noindent\textbf{VOT dataset.}
We validate the proposed tracker on the VOT-2015 dataset.
The dataset contains 60 short sequences with various challenges.
The VOT benchmark evaluates a tracker from two aspects: robustness and accuracy, which are different from the OTB benchmark.
The robustness of a tracker is measured by the failure times.
A failure is detected when the overlap ratio between the prediction and the ground truth becomes zero.
After 5 frames of the failure, the tracker is re-initialized to track the targets.
The accuracy of a tracker is measured by the average overlap ratio between the predicted results and the ground truths.
Based on these two metrics, Expected Average Overlap (EAO) is used for overall performance ranking.

Table~\ref{tab:VOT2015} shows the experimental results on the VOT-2015 dataset.
The proposed tracker performs favorably against the state-of-the-art trackers on this dataset.
We achieves the second-best EAO score (0.327) with the best accuracy (0.59)
and a favorable robustness score (1.09) close to the best one (0.95).
FlowTrack equipped with optical flow achieves the best EAO score (0.341). However, it runs at a slow speed (12 FPS) when compared to the proposed tracker (33.7 FPS).
For the VOT-2016 dataset, the proposed tracker obtains the best accuracy score (0.55) and the second-best EAO score (0.299).
Compared with the C-COT tracker, which achieves the best EAO score (0.331) and the best robustness (0.85), the proposed algorithm runs faster (33.7 vs. 0.3 FPS).
Overall, the proposed tracker performs well in terms of accuracy, robustness, and running speed.
It is worth noting that the favorable performance is achieved without an online update or offline training.
This demonstrates the effectiveness of the proposed deep features with target-active and scale-sensitive information, which helps to distinguish between the target objects and the background.

\vspace{1mm}
\noindent\textbf{Temple color-128 dataset.}
We report the results on the Temple color-128 dataset, which includes 128 color sequences and uses the AUC score as the evaluation metric.
Table~\ref{tab:Tcolor128} shows that the proposed algorithm achieves the best performance among the real-time trackers with an AUC score of 0.562.
The proposed tracker is not specially designed for these color sequences and does not exploit additional online adaption schemes, while it achieves a favorable performance and runs at real-time.
This shows the generalization ability of the proposed algorithm.

\begin{table}[!htb]
\small
\renewcommand\arraystretch{1.1}
\centering
\caption{\label{tab:Tcolor128} \textbf{Experimental results on the Temple color-128 dataset.} The notation (*) indicates the number is reported by the authors.}
\begin{tabular}{ l c c c}
\toprule
Method      	   & overlap-AUC 		& Real-time & FPS    \\
\midrule
MCPF~\cite{MCPF}   &   0.545			&N         &  1*   		\\
STRCF~\cite{STRCF} & 0.553 	 		    &N         &  6  		\\
C-COT~\cite{CCOT}  &  0.567 		    &N         &  1*	        \\
 MDNet~\cite{MDNET}& 0.590      		&N         &  1         \\
 ECO~\cite{ECO}    & 0.600  	    	&N         &  3    	 	\\
STRCF-deep~\cite{STRCF}  & 0.601 	  	&N         &  3  		\\
\midrule
STAPLE~\cite{staple}  &  0.498        	&Y          & 50          \\
BACF~\cite{BACF}        &  0.52  		& Y        &  35*     	 \\
ECO-HC~\cite{ECO}      & 0.552  	 	&Y         &  30     		\\
\textbf{Ours}       & \textbf{0.562} & \textbf{Y}  &  \textbf{33.7} \\
\bottomrule
\end{tabular}
\end{table}

\subsection{Ablation Studies}
In this section, we analyze the proposed method on the OTB datasets, including the OTB-2013 and OTB-2015 datasets, to study the contributions of different losses and different layer features.
Table~\ref{tab:ablation} presents the overlap ratio in terms of AUC scores of each variation.
The features from the output of the \textit{Conv4-3} and \textit{Conv4-1} layers are denoted as Conv4-3 and Conv4-1, respectively.
We compare the results of different feature layers based on regression loss, ranking loss, and random selection (randomly selecting the same number of filters), which are denoted as Regress, Rank, and Rand, respectively.
Compared with the random selection model, the regression loss scheme obtains significant gains in AUC scores for both the \textit{Conv4-1} (+4.3\% and +4.4\%)  and \textit{Conv4-3} (+4.9\% and +3.4\%) on the OTB-2013 and OTB-2015 datasets.
We attribute these gains to the benefits from the regression loss, which helps to select the most effective convolution filters to generate target-aware discriminative features.
By exploiting the objectness and semantic information pertaining to the target, the generated features are effective in distinguishing the target from the background and are robust to target variations.
The combination of regression-loss guided features from the \textit{Conv4-1} and \textit{Conv4-3} layers slightly improves the performance (+0.7\% and +0.7\%) on these two datasets.
This shows that although from different layers, these filters guided with the same loss do not provide much complementary information.
When combining different CNN layer guided by different losses, the improvement becomes larger (+1.8\% and +1.6\%).
The improvement benefits from the scale-sensitive information of the ranking-loss based features, which puts more emphasis on spatial details.
The comparison on the last two rows in Table~\ref{tab:ablation} demonstrates the effectiveness of the ranking loss.

\begin{table}[!htb]
	\small
    \renewcommand\arraystretch{1.1}
    \caption{\textbf{Ablation studies on the OTB dataset.}}
    \centering
    \begin{tabular}{lccccc}
    \toprule
    Conv4-1    & Conv4-3  & OTB-2013   & OTB-2015     \\
    \midrule
     Rand       &    --     &  0.602  & 0.597     \\
       --       &   Rand    &  0.618   & 0.610   \\
      Regress      &    --     &  0.645  & 0.646    \\
       --       &    Regress   &  0.662  & 0.644    \\
      Regress      &    Regress   &  0.669  & 0.651    \\
      Regress+Rank   &    Regress   &  0.680  & 0.660    \\
      \bottomrule
    \end{tabular}
    \label{tab:ablation}
\end{table}
\section{Conclusions}
In this paper, we propose to learn target-aware features to narrow the gap between pre-trained classification deep models and tracking targets of arbitrary forms.
Our key insight lies in that gradients induced by different losses indicate the importance of the corresponding filters in recognizing target objects.
Therefore, we propose to learn target-aware deep features with a regression loss and a ranking loss by selecting the most effective filters from pre-trained CNN layers.
We integrate the target-aware feature model with a Siamese tracking framework and demonstrate its effectiveness and efficiency for visual tracking.
In summary, we provide a novel way to handle the problems when using pre-trained high-dimensional deep features to represent tracking targets.
Extensive experimental results on five public datasets demonstrate that the proposed algorithm performs favorably against the state-of-the-art trackers.

\section*{Acknowledgments}
\small
This work is supported in part by the NSFC (No. 61672183), the NSF of Guangdong Province (No. 2015A030313544), the Shenzhen Research Council (No. JCYJ20170413104556946,
JCYJ20170815113552036, JCYJ20160226201453085), the Shenzhen Medical Biometrics Perception and Analysis Engineering Laboratory, the National Key Research and Development Program of China (2016YFB1001003), STCSM (18DZ1112300), the NSF CAREER Grant No.1149783, and gifts from Adobe, Verisk, and NEC.
Xin Li is supported by a scholarship from China Scholarship Council (CSC).

{\small
\bibliographystyle{ieee_fullname}
\bibliography{Tracking}
}

\end{document}